\documentclass{Interspeech}

\interspeechcameraready

\title{Adapting Foundation Speech Recognition Models to Impaired Speech:\\A Semantic Re-chaining Approach for Personalization of German Speech}

\author[affiliation={1,2}]{Niclas}{Pokel}
\author[affiliation={2}]{Pehuén}{Moure\,$^*$}
\author[affiliation={2}]{Roman}{Boehringer\,$^*$}
\author[affiliation={3}]{Yingqiang}{Gao\,$^*$}

\affiliation{School of Computation, Information and Technology}{Technical University of Munich}{Germany}
\affiliation{Institute of Neuroinformatics}{University of Zurich and ETH Zurich}{Switzerland}
\affiliation{Department of Computational Linguistics}{University of Zurich}{Switzerland}
\email{niclas.pokel@tum.de, pehuen@ini.ethz.ch, roman@ini.ethz.ch, yingqiang.gao@cl.uzh.ch}
\keywords{automatic speech recognition, dysarthric speech, impaired speech}

\newcommand{\peh}[1]
{{\color{teal}\bgroup\textbf{[peh:}}~#1\textbf{]}\egroup}

\usepackage{comment}
\usepackage{booktabs}
\usepackage{multirow}
\usepackage{multicol}
\usepackage{hyperref}
\usepackage[dvipsnames]{xcolor}

\begin{document}

\maketitle

\renewcommand{\thefootnote}{\fnsymbol{footnote}}
\footnotetext[1]{Contributed equally as co-last authors.}
\renewcommand{\thefootnote}{\arabic{footnote}}

\begin{abstract}

 Speech impairments caused by conditions such as cerebral palsy or genetic disorders pose significant challenges for automatic speech recognition (ASR) systems. Despite recent advances, ASR models like Whisper struggle with non-normative speech due to limited training data and the difficulty of collecting and annotating non-normative speech samples. In this work, we propose a practical and lightweight pipeline to personalize ASR models, formalizing the selection of words and enriching a small, speech-impaired dataset with semantic coherence. Applied to data from a child with a structural speech impairment, our approach shows promising improvements in transcription quality, demonstrating the potential to reduce communication barriers for individuals with atypical speech patterns.

\end{abstract}


\section{Introduction}

Recent advances in deep learning have revolutionized automatic speech recognition (ASR), with large foundational models such as Whisper \cite{radford2022whisper} and Wav2vec \cite{schneider2019wav2vec}  achieving near-human performance on normative speech. However, these models face significant challenges when processing impaired speech, such as dysarthria speech disorder \cite{jayaraman2023dysarthria}. 
Although individuals with speech impairments typically maintain intact language comprehension and cognitive abilities, their decreased speech intelligibility leads to reduced communication and integration \cite{page2022communicative, van2023automatic, troger2024automatic}.

Deep learning models for ASR struggle with impaired speech primarily due to the limited availability of training data and the fundamental deviations of such speech from normative patterns and inter-individual variability\cite{baskar2022speaker, hermann2023few, tobin2024automatic}. Given the considerable variability in speech characteristics among individuals with speech impairments, personalization of ASR models for a specific user's data is crucial to effectively addressing these differences \cite{shor2019personalizing}. 
This challenge is particularly acute for language-specific models, such as those for German-impaired speech, where data-scarcity poses an additional barrier.
Consequently, there is growing research interest in enhancing the performance of ASR models for German impaired speech with minimal reliance on extensive data augmentation.


Many works of deep learning based dysarthric or general impaired speech recognition have introduced a range of data augmentation strategies, including diffusion-based text-to-speech (TTS) synthesis \cite{leung2024training}, few-shot learning with multi-speaker TTS systems \cite{hermann2023few}, two-stage augmentation pipelines combining phonetic perturbations with dysarthria-specific transformations \cite{bhat2025two}, and severity-aware synthetic speech generation \cite{geng2023use, soleymanpour2024accurate}. While these methods have demonstrated promising improvements, they are often developed and evaluated on sentence-level utterances that do not align well with widely used corpora such as UA-Speech \cite{kim2008dysarthric}, which consists of isolated word-level utterances.

Moreover, collecting and annotating sentence-level data often necessitates the involvement of individuals familiar with the speaker’s unique speech patterns, typically close relatives or trained caregivers \cite{article2006}. This reliance, coupled with the complexity of longer utterances, poses significant challenges for data acquisition and annotation outside of clinical or highly supported environments. Consequently, deploying personalized ASR systems in everyday, private settings remains difficult.



Our core \textbf{contributions} to address the above challenges are:
\begin{enumerate}
    \item The creation of BF-Sprache (``\textit{Barrierefreie-Sprache}'' (\textit{barrier-free speech}) in German) the first German impaired speech dataset containing 505 words with broader biphone coverage than existing resources, designed to mirror the principles of UA-Speech while filling a critical gap for German language resources;
    \item Two novel word selection algorithms: Greedy Biphone Coverage (GBC) for maximizing phonetic diversity and Personalized Weighted Phoneme Selection (PWPS) for targeting clinically significant phonemes;
    \item A semantic re-chaining (SRC) pipeline that constructs coherent sentence-level utterances from isolated word recordings, enabling ASR models to better leverage their internal linguistic representations while minimizing data collection burden;
    \item Comprehensive experimental evaluation demonstrating significant error rate reductions across varying intelligibility levels for both UA-Speech and BF-Sprache in both zero-shot and fine-tuned models
\end{enumerate}

Our approach relies on simple, easily collected data, enabling personalization in everyday environments and marking an important step toward more accessible speech technology for individuals with speech impairments.

\section{Dataset}


Following the design principles of the UA-Speech dataset, we introduce BF-Sprache, a novel dataset designed to personalize ASR models for German non-normative speech. Below, we briefly summarize the two datasets used in this study.

\begin{table*}[t!]
    \centering
    \caption{CER (\%) results of zero-shot and fine-tuning on different data splitting settings. }
    \resizebox{0.7\linewidth}{!}{
    \begin{tabular}{lcccccc}
    \hline
     \textbf{Method} & \textbf{Intelligibility} & \textbf{Single Word}  & \textbf{Strict} & \textbf{Mixed} & \textbf{Natural} & \textbf{Mixed + SRC}  \\
     \hline
     \multirow{3}{*}{Zero-shot} & Very Low & 100.30 & 92.47 & 90.81 & \underline{90.62} & \textbf{83.97} \\
     & Low & 84.91 & 68.29 & 68.96 & \underline{65.25} & \textbf{53.91} \\
     & Medium & 68.36 & 39.29 & 38.91 & \underline{37.22} & \textbf{33.15} \\
     \hline
     \multirow{3}{*}{Fine-tuning} & Very Low & 60.35 & 72.23 & \underline{42.66} & 43.52 & \textbf{23.05} \\
     & Low & 24.40 & 34.42 & 19.01 & \underline{18.61} & \textbf{16.64} \\
     & Medium & 19.93 & 14.71 & 12.92 & \underline{12.86} & \textbf{6.62} \\
     \hline
    \end{tabular}
    }
    \label{tab:0shot_train}
\end{table*}

\subsection{The UA-Speech Dataset}

The UA-Speech dataset \cite{kim2008dysarthric} is one of the most widely used benchmarks in dysarthric speech recognition research and has laid foundation for many deep learning based works \cite{schu2023using, shih2022dysarthria, fernandez2020attention, cadet2024study}. It consists of recordings from 19 speakers with varying degrees of dysarthria, as well as 10 control speakers. The dataset emphasizes isolated word-level utterances, comprising 455 distinct words including digits, letters, and phonetically rich uncommon words. Each of the 155 common words was spoken three times by every participant. The 300 phonetically rich uncommon words on the other hand, were spoken once. Each utterance was recorded over 7 microphones. This structured design enables controlled experimentation and fine-grained analysis of ASR performance across different severity levels of speech impairment. Despite its utility, UA-Speech presents two key limitations: (1) it lacks the semantic coherence needed to effectively personalize ASR models for naturalistic speech, and (2) it is limited to English, thereby restricting its applicability to non-English ASR systems.



\subsection{The BF-Sprache Dataset}

We construct the BF-Sprache dataset by first selecting German words from a large-scale German corpus (e.g., Project Gutenberg \cite{projectgutenberg}). Subsequently, word-level utterances are recorded from a speech impaired native German speaker to create the final dataset. The BF-Sprache dataset serves dual purposes: (1) it can be used to personalize ASR models for transcribing German impaired speech, and (2) it offers potential as a resource for generating materials to support clinical practices in logopedics. We propose the following two algorithms for selecting German words to be used as targets for utterance recording:

\subsubsection{Greedy Biphone Coverage (GBC)}

With a focus on enriching phoneme diversity, this algorithm aims at maximizing the coverage of unique biphones (i.e., pairs of consecutive phonemes) under a fixed word budget $k$. 


Let $W$ be the set of candidate words, and \( B(w) \) the set of biphones for a given word \( w \in W \). 
The GBC algorithm iteratively adds a word \( w^*_i \) to the current selected set $S_i$ ($i$ ranges from 1 to $k$) to maximize the number of new biphones introduced at each step $i$:
\begin{align*}
    w^*_i = \mathop{\arg\max}_{w \in \{W \setminus S_{i-1}\}} 
\bigl\lvert B(w) \setminus {\bigcup} \hspace{0.2em} B(w'), \forall w' \in S_{i-1}  \bigr\rvert,
\end{align*}
where \( S_{i-1} = \{w^*_1, \dots, w^*_{i-1}\} \) is the set of selected words at last step (with initially \( S_0 = \varnothing \)). The selection process terminates when the budget \( k \) is reached or no candidate word introduces new biphones.

\subsubsection{Personalized Weighted Phoneme Selection (PWPS)}

After selecting a base set of words with GBC, this algorithm further selects an additional $k'$ words, prioritizing a specific target set of phonemes $P_t$ associated with a weight $\alpha_p$ for each $p \in P_t$. This set of target phonemes can be, for instance, derived from a logopedic report of the speech-impaired individual.

Let $W'$ denote the set of remaining words not selected with GBC (i.e., words not present in $S_k$), and $P(w)$ the set of phonemes for a given word $w \in W'$. The PWPS algorithm aims to improve the coverage of phonemes deemed important in the individual's logopedic assessment report. At each step $j$ ($j$ ranges from 1 to $k'$), PWPS determines a word $w_j^*$ by performing
\[
w'^*_j = \underset{w \in \{W' \setminus S'_{j-1}\}}{\arg\max} \sum_{p \in \{ P(w) \cap P_t \} } \frac{\alpha_p}{c(S'_{j-1}, p) + 1}
\]
where $S'_{j-1} = \{w'^*_1, \dots, w'^*_{j-1}\}$ is the set of previously selected words (similarly, $S'_0 = \varnothing$) and $c(S'_{j-1}, p)$ counts the occurrences of phoneme $p$ across all words in $S'_{j-1}$ (we applied Laplace smoothing to avoid numerical issues). 
The weight $\alpha_p$ for each phoneme is empirically derived from the individual's logopedic report to reflect its therapeutic importance.
While GBC ensures general phoneme diversity in the BF-Sprache dataset, the PWPS stage further prioritizes phonemes that are clinical significant but underrepresented, thereby enhancing the dataset's utility for personalizing impaired speech models. 

Table~\ref{tab:statistics} summarizes the core comparison of the two datasets of this work. After this two-stage word selection process, we gathered 505 German words in total and recorded their utterances with one speech-impaired participant (under adulterant female, Swiss-German native with Apert-Syndrome). We set up a standard recording setup with one channel\footnote{Microphone: Rode NT1-A via XLR, Interface: Focusrite Scarlett 2i2, Software: Audacity v3.7.3}. 
All recordings were conducted in a single session, with short breaks provided for the participant. 
Following this, semantically coherent sentences (Original Sentences) as well as randomly generated word sequences (Random Sentences), constructed from the previously recorded individual words, were also recorded. 
This was done to evaluate the model’s performance on natural speech artifacts such as pauses, prolongations, and fillers, and to assess its generalization from low-artifact scenarios (e.g., short, isolated read phrases) to high-artifact scenarios (e.g., free speech or longer read sentences).
We used a universal multilingual phoneme recognizer Allosaurus \cite{li2020universal} in this work.

Prior to the collection of data, we have obtained the approval from the university's ethics committee which allows us to construct and publish the data. 


\begin{table}
    \centering
    \caption{Statistics of the two datasets used in this study. \# Words refers to the total number of words, including both common and uncommon entries, while \# Biphones indicates the total number of distinct subsequent phonemes covered.}
    \begin{tabular}{llcc}
    \hline
      \textbf{Dataset} & \textbf{Usage} & \# \textbf{Words} & \# \textbf{Biphones}  \\
      \hline
      UA-Speech & Generalization & 455 & 337\\
      BF-Sprache & Personalization & 505 & 731\\
      \hline
\end{tabular}
    
    
    \label{tab:statistics}
\end{table}

\section{Methods}


To bridge the gap between isolated word-level utterances and the semantic coherent naturalistic speech, we introduce a \textbf{semantic re-chaining} (SRC) approach that assembles sentence-level utterances from existing word-level recordings.

Given a set of $N$ incoherent, isolated word-level utterances $(u_i)_{i=1}^N$, our goal is to identify a subset of utterances and combine them into a new sentence-level utterance $\tilde{u}$
\begin{align*}
    \tilde{u} = f\left[(u_j)_{j=1}^m \sim (u_i)_{i=1}^N\right], \text{ with } 1 \leq j \leq m \leq N,
\end{align*}
where $f$ is a mapping function that concatenates $(u_j)_{j=1}^m$ to $\tilde{u}$ with a specific order. A semantically coherent $\tilde{u}$ would therefore requires $f$ to be, for example, a pre-trained language model that generates texts of low perplexity, or a human speech therapist who prescribes sentences for the speech-impaired individual to practice. Mathematically, 
\[
f\left[(u_j)_{j=1}^m\right] =
\begin{cases}
\texttt{LM}\left[(u_j)_{j=1}^m\right], & \text{language model or} \\
\texttt{Manual}\left[(u_j)_{j=1}^m\right], & \text{speech therapist.} \\
\end{cases}
\]

\begin{figure*}[htb]
    \centering
    \includegraphics[width=0.8\linewidth]{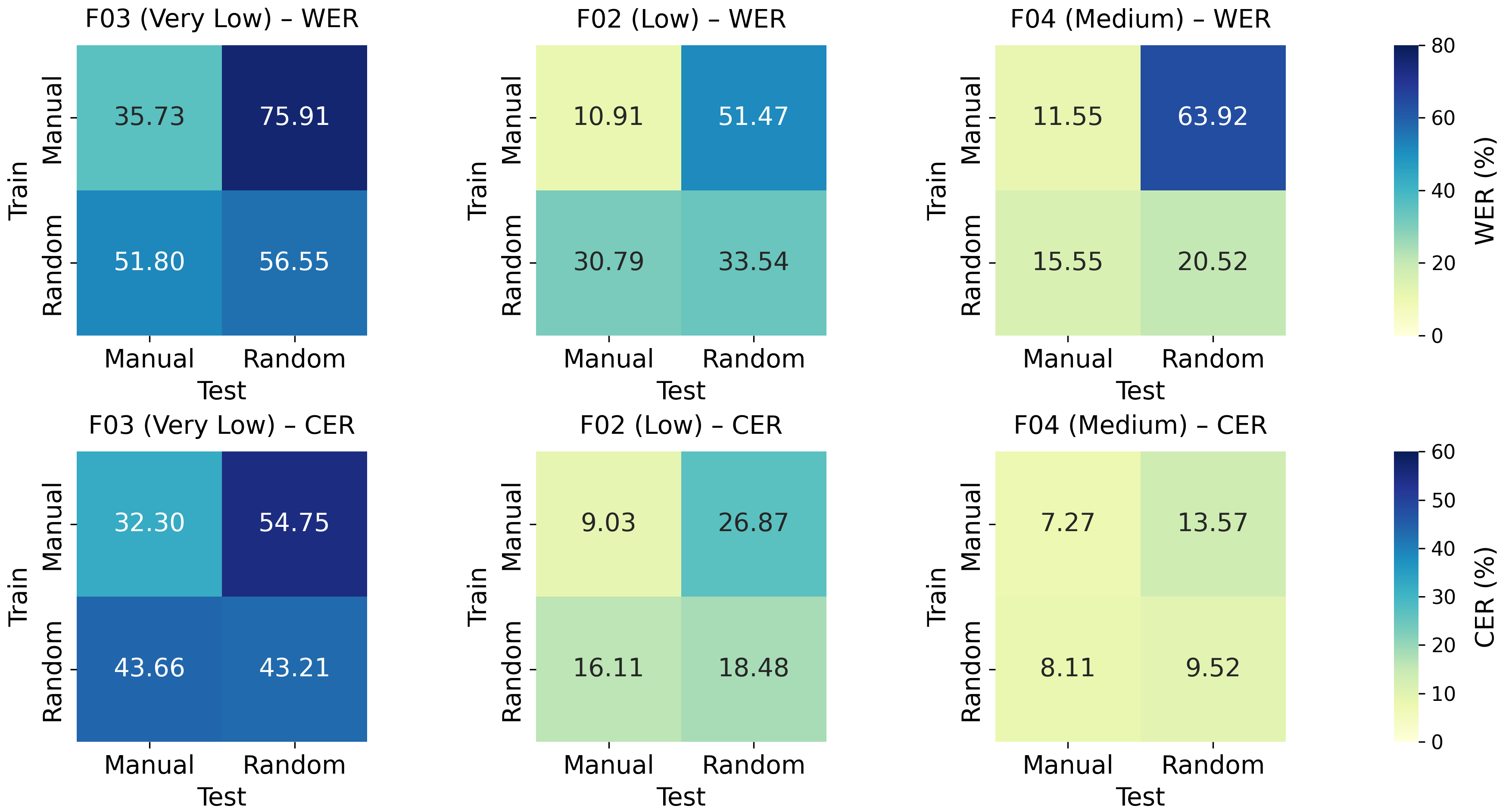}
    \caption{Word error rate (WER in \%, top row) and character error rate (CER in \%, bottom row) for three UA‑Speech speakers with differing intelligibility levels—F03 (Very Low), F02 (Low), and F04 (Medium) and with "Natural" split. Each 2×2 heat‑map shows the effect of training the dysarthric ASR model with versus without manual semantic re‑chaining data (Train: Manual vs. Random) and evaluating with versus without re‑chaining in the test set (Test: Manual vs. Random).}
    \label{fig:enter-label}
\end{figure*}

As recording and annotating naturalistic impaired speech is often time-consuming and labor-intensive, our semantic re-chaining approach offers an efficient alternative by synthesizing new training data from pre-recorded word-level utterances, that are easy to record and annotate. 

In this work, we initially asked a native German speaker to produce semantically coherent sentences. However, due to the time-intensive nature of this task, we later outsourced the task to commercial language models, namely, Gemini 2.5 Pro \cite{Kavukcuoglu2025Gemini25} and GPT-o3 mini \cite{openai2025o3mini}. The generated sentences were subsequently checked for inappropriate content and manually validated for semantic coherence and meaningfulness.



In order to test if semantic information is crucial for the ASR model to precisely recognize impaired speech, we also conduct experiments using sentence-level utterances constructed without preserving semantic meaning. This is achieved by randomly concatenating word-level utterances from a candidate pool with bootstrapping.

Let $U = (u_i)_{i=1}^N$ be the fixed set of $N$ available word-level utterances. To generate a random sentence-level utterance of length $m$, we select $m$ words by drawing independently and uniformly from $U$. Let the sequence of selected words be $(w_j)_{j=1}^m$. Each word $w_j$ in this sequence is chosen independently and uniformly from $U$.
Formally, for a generated sequence of words $(w_j)_{j=1}^m$:
\[
w_j \stackrel{iid}{\sim} \mathcal{U}(U) \quad \text{for } j=1, \dots, m
\]
where $\mathcal{U}(U)$ denotes the uniform distribution over the word-level utterances in $U=(u_i)_{i=1}^N$. 
The final random sentence-level utterance is the concatenation of these words.

 
\section{Experimental Designs}

To test the effectiveness of our personalization pipeline, we set up the following train-test split routines: 

\begin{itemize}
    \item \textbf{Strict (Vocabulary-Based Split):} Groups all instances of the same word together, ensuring zero lexical overlap by design. However, this approach is non-standard for typical ASR tasks and likely underestimates generalization.
    \item \textbf{Mixed (Session/Repetition Grouped Split):} Groups all microphone recordings from the same speaker and repetition block. Entire blocks are randomly assigned to either train or test, preventing microphone leakage within a session while allowing realistic lexical overlap across different sessions.
    \item \textbf{Natural (Word-Repetition Split):} Groups recordings by (Speaker, Session), but assigns the individual per-word blocks recordings to train or test. This prevents leakage and introducing more diverse combinations compared to Mixed.
\end{itemize}


Among all split settings, Natural and Mixed better reflect real-world conditions, despite not guaranteeing fully disjoint train-test sets. They ensure that words can form coherent sentences, unlike Strict splitting, where models may struggle to generalize to realistic data. We still include Strict to assess the model’s upper capacity.

For all experiments, we used Whisper-Large V3 \cite{radford2022whisper} as the backbone of the impaired speech model, evaluating its performance under both zero-shot and supervised fine-tuning settings, with and without augmentation via semantic re-chaining. We used word error rate (WER) and character error rate (CER) as metrics\footnote{We used the Python package \texttt{\href{https://github.com/jitsi/jiwer}{jiwer}}, Apache 2.0 License.} to evaluate our experiment outcomes. 

To evaluate real-world performance and generalization, we reused previously recorded utterances from the speech-impaired participant, including short story readings and free conversations (with parental speech removed during processing). These recordings contained idiosyncratic, region-specific terms, such as local place names, posing challenges for ASR models that are unlikely pre-trained on such vocabulary.

\section{Results and Discussion}

\subsection{Performance on UA-Speech}

\begin{table}[t!]
    \centering
    \caption{Zero-shot CER (\%) for single speaker training with and without the data augmentation of semantic re-chaining (SRC). * in the table indicates semantic structure being removed and not artificially added, since the data contains full read sentences.}
    \resizebox{\columnwidth}{!}{
    \begin{tabular}{lccc}
    \hline
     \textbf{Dataset} & \textbf{Intelligibility} & \textbf{No SRC} & \textbf{With SRC}  \\
     \hline
     \multirow{4}{*}{UA-Speech} & Very Low & 90.62 & \multicolumn{1}{l}{\textbf{83.97} (\textcolor{ForestGreen}{7.34 \%}  $\downarrow$)} \\ 
     & Low & 68.96 & \textbf{53.91} (\textcolor{ForestGreen}{21.82 \%}  $\downarrow$) \\ 
     & Medium & 39.91 & \textbf{33.15} (\textcolor{ForestGreen}{16.94 \%} $\downarrow$) \\ 
     & High & 39.70 & \textbf{20.84} (\textcolor{ForestGreen}{47.50 \%} $\downarrow$) \\

    \hline
     BF-Sprache & Low & 87.93* & \textbf{39.28} (\textcolor{ForestGreen}{\textbf{55.33} \%} $\downarrow$) \\
     \hline
    \end{tabular}
    }  
    \label{tab:zero-shot-CER}
    
    \vspace{1mm}
    \small
    \raggedright
    
\end{table}

Table~\ref{tab:zero-shot-CER} shows the results of zero-shot CER measured on data of a single speaker in both datasets. 

We observed that SRC had a substantial impact on both word and character error, as shown in Table~\ref{tab:0shot_train}. 
Across experiments and random seeds, SRC yielded the largest relative improvements for speakers with either medium or very low intelligibility for finetuning. 
A plausible hypothesis is that for speakers with low intelligibility, the model increasingly relies on its semantic prior knowledge, effectively ``inferring'' more, whereas high-intelligibility speakers are transcribed so accurately that Whisper's autoregressive structure can smooth over minor semantic inconsistencies.

Interestingly, in the zero-shot setting, speakers with the highest intelligibility benefited most. This aligns with the previous hypothesis: for speakers with very low intelligibility, the transcriptions are nearly unusable in zero-shot scenarios, rendering Whisper's semantic priors largely ineffective.

These observations remained consistent and intuitively plausible even when training and test sets were processed independently. Across all configurations, applying SRC to both training and test sets yielded the best performance, as shown by Figure~\ref{fig:enter-label}. 
These effects were more pronounced in the WER than in the CER, indicating that basic transcription and intelligibility are already functioning well. Semantic adaptation appears to primarily refine near-correct predictions by resolving subtle contextual inconsistencies.

\subsection{Performance on BF-Sprache}

\begin{table}[htb]
\centering
\caption{Fine-tuned CER (\%) for different configurations.}
\resizebox{0.7\linewidth}{!}{
\begin{tabular}{l c}
\toprule
\textbf{Configuration} & \textbf{CER (\%)} \\
\midrule
Baseline (BL) & 23.38 \\
BL + Original Sentences & 19.89 \\
BL + Artificial Sentences & 20.33 \\
BL + Random Words & 22.50 \\
BL + All & 21.39 \\
\midrule
Original Sentences & 28.02 \\
Artificial Sentences & 28.19 \\
Random Words & 38.89 \\
\bottomrule
\end{tabular}
}

\label{tab:BF_configs}
\end{table}

Table~\ref{tab:BF_configs} shows our results on the BF dataset. Baseline (BL) refers to training on read and one split of free speech, while testing on the other free speech data. 

Overall, we observed similar results on BF-Sprache and UA-Speech: semantically structured training sets led to significantly better generalization to the target domain than randomly concatenated word sequences.

Interestingly, model performance was not meaningfully impacted by whether sentences were recorded naturally or synthetically composed from individual word recordings. This finding supports two possible hypotheses: either the autoregressive nature of Whisper models effectively smooths over artifacts such as fillers and extended pauses, or the artifact density in read speech remains too low to enable robust generalization to spontaneous speech. The former is supported by the near-identical performance between real and synthetic sentences, while the latter is suggested by the model’s still moderate performance on naturalistic speech.

A central limitation at this stage is the small dataset size, which makes it difficult to conclusively disentangle these effects. In future work, we plan to expand the dataset to include data from multiple individuals with speech impairments.

\section{Conclusions and Future Work}

In this work, we propose a lightweight pipeline for personalizing ASR models for a German native speech-impaired individual. We introduce BF-Sprache—a novel speech dataset of word-level utterances recorded for German non-normative speeches. Through practical data augmentation using two proposed algorithms, we construct semantically coherent sentences that better reflect naturalistic speech. Our approach relies on simple, easily collected data, enabling personalization in everyday, non-clinical settings. Experiments show that, even with limited German non-normative data, our pipeline significantly reduces transcription errors with minimal human intervention, marking a step toward more accessible speech technology for individuals with speech impairments.

We are currently expanding the BF-Sprache dataset to include a broader range of target demographic groups and to refine our data collection methodology with greater expert involvement, for example, by consulting with speech therapists and incorporating clinical logopedic reports of individuals with speech impairments.  Our overarching goal is to develop personalized ASR models that adapt to each individual with speech impairments, using training data that can be easily and comfortably acquired at home. Our approach aims to enable more accurate and accessible speech recognition, tailored to each person’s unique speech patterns and communication needs.

\cleardoublepage 

\section{Acknowledgment}

This work is partially based on previous work from Philipp Guldimann, who developed the initial codebase and collected and annotated portions of the audio data used \cite{Guldimann2024}. 

\bibliographystyle{IEEEtran.bst}
\bibliography{mybib.bib}


\end{document}